\documentclass[runningheads]{llncs}
\usepackage{graphicx}
\RequirePackage[hyphens]{url}
\usepackage{booktabs}
\usepackage{amsmath}
\usepackage{caption}
\usepackage{subcaption}
\begin{document}
\title{MusIAC: An extensible generative framework for Music Infilling Applications with multi-level Control}
\titlerunning{MusIAC: Music Infilling Applications with multi-level Control}
%
\author{Rui Guo\inst{1}\orcidID{0000-0003-2907-0718} \and
Ivor Simpson\inst{2}\orcidID{0000-0001-5605-6626} \and 
Chris Kiefer\inst{1}\orcidID{0000-0002-3329-1938} \and 
Thor Magnusson\inst{1}\orcidID{0000-0002-3731-0630} \and
Dorien Herremans\inst{3}\orcidID{0000-0001-8607-1640}}
\authorrunning{Rui Guo et al.}
%
\institute{Department of Music, University of Sussex
\email{r.guo@sussex.ac.uk}\and
Department of Informatics, University of Sussex \and
Information Systems Technology and Design, Singapore University of Technology and Design
}
\maketitle              
\begin{abstract}

We present a novel music generation framework for music infilling, with a user friendly interface. Infilling refers to the task of generating musical sections given the surrounding multi-track music. The proposed transformer-based framework is extensible for new control tokens as the added music control tokens such as tonal tension per bar and track polyphony level in this work. We explore the effects of including several musically meaningful control tokens, and evaluate the results using objective metrics related to pitch and rhythm. Our results demonstrate that adding additional control tokens helps to generate music with stronger stylistic similarities to the original music. It also provides the user with more control to change properties like the music texture and tonal tension in each bar compared to previous research which only provided  control for track density. We present the model in a Google Colab notebook to enable interactive generation.

\end{abstract}
\section{Introduction}
Music composition by artificial intelligence (AI) methods, especially using deep learning, has been an active topic of research in recent years \cite{briot2020deep,jiComprehensiveSurveyDeep2020}. In a recent survey of musical agents \cite{tatarMusicalAgentsTypology2019}, twelve musical tasks such as accompaniment generation, melody/rhythm generation, continuation, arrangement, and style imitation are examined. In the deep learning era, all of these tasks have been explored to some extent. 

When applying AI to music composition, however, an often ignored question is ``why'' one might wish computers to compose music. From the perspective of the deep learning practitioner, the answer may be to explore the limits of AI models for creative tasks, and investigate whether they can generate music as human composers. On the other hand, musicians and composers may want to use AI as a source of inspiration, for instance, by rapidly offering several solutions. One such AI method is music infilling or inpainting~\cite{patiLearningTraverseLatent2019a,hadjeres2021piano}. It is used to extend pre-existent music materials, such as filling in the missing bars or tracks given the surrounding music information. It can write a new melody line given the existing bass and accompaniment track, or rewrite a few bars in the middle given the beginning and the end.  Many reasonable solutions may exist that match the surrounding music progression and harmony. Without efficient track and bar music property conditions, however, the user has to generate the music repeatedly until it satisfies user's requirement.

Several research studies have used a transformer model \cite{vaswani2017attention} for symbolic music generation \cite{huang2018music,huang2020pop,renPopMAGPopMusic2020,ensMMMExploringConditional2020,hsiaoCompoundWordTransformer2021,hadjeres2021piano} and the results are promising. However, controlling the generation process is still limited in these approaches.

One common control for the music infilling system is track density\cite{ensMMMExploringConditional2020,hadjeres2021piano}, which is defined as the number of notes in a track divided by the total timesteps in that track. However, a sole density cannot easily change the accompaniment track from a monophonic style to a polyphonic style. A polyphony control can help to convert a monophonic track such as arpeggio to a more polyphonic texture such as a chord track or vice versa in a direct way, and that can be useful mostly for the accompaniment track. Another interesting control is the track occupation rate, which determines which ratio of a track is note occupied versus filled with rests. These track features may be useful as a composer may want to control the track texture.

Except for those track controls, a bar level tonal tension control\cite{chewSpiralArrayAlgorithm2002,herremansTensionRibbonsQuantifying} can help to create music with specific tension movements, e.g. from low to high, high to low or any tension shape. One use case is to change the tension of the beginning and ending of a piece so as to set certain moods.

To implement these controls, the respective track/bar properties are calculated and added to the input. We deliberately choose to use higher level human interpretable parameters as controls, including six features: key, bar tensile strain, bar cloud diameter, track density, track polyphony, and track occupation, and they are calculated from the input sequence directly. It may be useful to generate music according to the track/bar control parameter template fit to a particular scenario, such as high track note density, low track polyphony rate and high track occupation. As the model learns the relationship between these control tokens and the music, the controls can be changed to generate variations of the original music. In the experiments, we observe that an additional benefit of including more music properties in the input is that the generated music is more similar to the original music measured by pitch and rhythm related metrics.

In this paper, we propose an extensible framework for music generation by infilling reconstruction. Six musically meaningful control tokens are calculated and added to the original input. The effect of adding this conditioning information is examined in an experiment that uses seven objective metrics selected from the literature. Our simple model design makes it extensible so that we can easily include additional tokens in the future. The music infilling task, which involves reconstructing a missing segment of music, is used to validate our results by comparing the properties of original and generated examples. The results show that the model with added calculated music tokens to the input has more stylistic similarity to the original music.  Google Colab notebook is shared for free exploration of this infilling system and gives a straightforward way to explore the effect of adding novel musically meaningful tokens to the input. The music generated by changing control tokens demonstrates the controllability of this method.

\section{Related Work}

Over the years, many different generative models have been developed for symbolic music generation \cite{jiComprehensiveSurveyDeep2020,herremans2017morpheus}. Variational AutoEncoder(VAE) based models \cite{tanMusicFaderNetsControllable2020,patiLearningTraverseLatent2019a,guo2020variational} usually generate short music pieces and explore different music features in the latent space. Generative Adversarial Network (GAN)  based models \cite{brunnerSymbolicMusicGenre2018a} can generate longer music, but can be harder to train and may suffer mode collapse without careful parameter tuning \cite{muhamedtransformer}. Recursive Neural Networks \cite{zixun2021hierarchical}, and more recently the powerful transformer based methods\cite{huang2018music} can generate long music pieces but with less control explored so far compared to the VAE models.

Several improvements have been made since the transformer model was first used for music generation, related to both the input representation and the model structure. \cite{huang2020pop} uses ``position''(timestep) and ``duration'' tokens to replace the ``note on'' and ``note off'' tokens \cite{oore2020time}. This allows the model to learn to pair the ``note on'' and ``note off'' if they are far apart. \cite{renPopMAGPopMusic2020} generates accompaniment given the melody track, and adds pitch, velocity, and duration embeddings in one timestep. \cite{hsiaoCompoundWordTransformer2021} has a similar design and  uses different output linear layers for different token types. The models by \cite{ensMMMExploringConditional2020,hadjeres2021piano} generate music infillings similar to the task tackled in this research. Both models take the track note density as the control parameter, without providing any other track/bar level control features, we will explore adding the latter features in this research. 

Some interactive interfaces have previously been designed specifically for the music infilling task. \cite{bazinNONOTOModelagnosticWeb2019} and \cite{louieCococoAISteeringTools2020}'s interfaces are based on music chorale generation \cite{huang2019counterpoint}.

\section{Proposed model and representation}

The existing transformer-based methods offer few semantic controls for the generation process, or focus on continuous generation rather than infilling. Given the right input controls, music generation models may be driven and steered by relevant musical concepts and ideas. Our work is based on the following assumptions:

\begin{enumerate}
  \item Including additional derived musical features can improve the performance of music infilling.
  
  \item Using human interpretable music features allows the user to control the generated music.
  
\end{enumerate}

\begin{figure}
\includegraphics[width=0.8\textwidth]{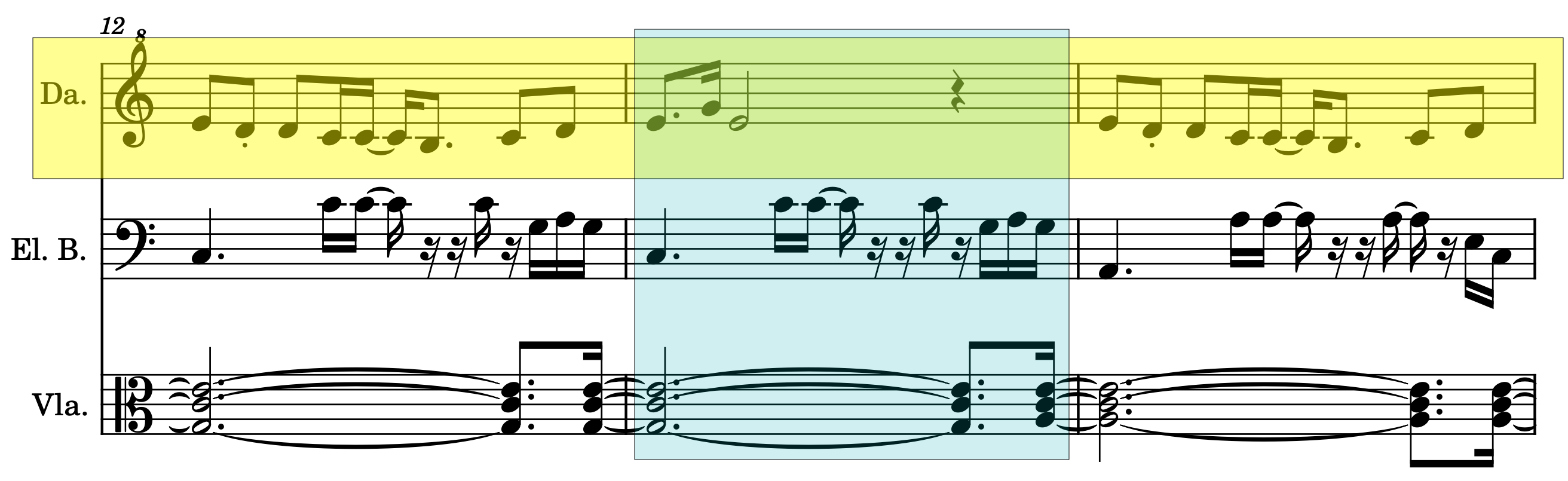}
\caption{An example of original music with two infilled regions. The yellow block masks the melody track, and the blue block masks the second bar. The notes of those two masked regions are replaced by a ``mask'' token in the model input. The target output of the model is to reconstruct the missing track/bar in the infilled region.} 
\label{infilling}
\end{figure}

Because the music infilling region is the model's prediction objective, it is natural to compare the generated music to the original. If the generated music has similar properties to the original infilled music region, then the model has performed well. Our model is versatile enough to allow multiple types of infilling. For instance, in pop music with multiple tracks, the infilling can work either by infilling a whole track or by infilling a bar across tracks, or both at the same time. \figurename~\ref{infilling} shows an example of how we can formulate the infilling task. The input music has three tracks, the yellow block region masks the first track, and the blue block region masks the second bar. The aim of the model here is to reconstruct the masked music region given the surrounding information. Providing input with multiple tracks makes it possible to have the track properties separately, and the control for different tracks can be tested separately.

\subsection{Adding control features}

We selected the following information to be added to the model input as controls from multiple levels. This is calculated from the MIDI data and provides high level musical concepts as conditions for the music generation.

\begin{enumerate}
   \item Track level controls:
   \begin{itemize}
     \item The track's note density rate: $number_{note} / timesteps_{total}$. This is calculated by dividing the number of notes in a track by the maximum time steps in that track. 
     
     \item The track's note polyphony rate: $timesteps_{poly note} / timesteps_{any note}$. This is the number of timesteps with more than two notes divided by the total number of timesteps with any note.
      
      \item The track's note occupation rate: $timesteps_{any note} / timesteps_{total}$. This is the total number of timesteps with any note divided by the total number of time steps, including those with rests.

   \end{itemize}
   
   \item Bar level controls:
   \begin{itemize}
     \item The tensile strain \cite{herremansTensionRibbonsQuantifying} of the notes in that bar: 
     $\sum_{i=1}^{n} (note_{pos}[i] - key_{pos}) / n $, which is the average of the difference from the note position to the key position. The note and key position are calculated based on the spiral array theory~\cite{chewSpiralArrayAlgorithm2002}. This is a tonal tension measure.
     
     \item The cloud diameter \cite{herremansTensionRibbonsQuantifying} of the notes in that bar: 
     
     $\max_{i \in [1..n-1],j \in [i+1..n]} (note_{pos}[i] - note_{pos}[j])$. This is another tonal tension measure, which only calculates the largest distance between notes in that bar. The calculation of the note position is also based on the spiral array theory.
   \end{itemize}
  
\end{enumerate}

Except for the above controls, the following information is also added to the model's input as auxiliary information. The key is calculated by \cite{cuthbert2010music21,guoMIDIMinerPython2019}. The tempo, time signature, and track instrument number are extracted directly from the MIDI files.

   \begin{itemize}
   \item The key of the song, which can be one of 24 keys (major and minor). 
     \item The tempo of the song, categorised into seven different bins.
     \item The time signature of the song, including 4/4, 3/4, 2/4, and 6/8.
     \item The track's instrument: The MIDI instrument number.

   \end{itemize}

\subsection{Data representation}

We use an adapted version of the ``REMI'' \cite{huang2020pop} token representation in this work. The ``REMI'' format includes position tokens to mark the position of the note inside a bar. The number of the position is related to the minimum duration of the note. We select the 16th note as the minimum note length, and a bar in 4/4 metre is thus divided into 16 different start positions. The position tokens range from ``e\_0''  to ``e\_15'', and the duration tokens range from ``n\_1'' to ``n\_32''. The maximum note length ``n\_32'' represents two whole notes in the time signature of 4/4.  The pitch tokens range from ``p\_21'' to ``p\_108'', representing A-1 to C7 respectively. There is a ``bar'' token to mark the start of a new bar. The velocity, tempo, and chord tokens proposed in \cite{huang2020pop} are discarded in the format used here. The dynamics of music is not the focus of this research, and by removing the velocity of each note, notes with the same duration can be grouped by using only one duration token after the pitch tokens. E.g. e\_0, p\_60, p\_67, n\_10 means note C3 and G3 have the same duration ($10\times16$th note), which equals the summation of a half note (8*16th note) and an eighth note ($2\times16$th note). Because the tonal tension information is included, the chord information is also removed. 

To represent the ``track'' concept, a ``track'' token is added to the vocabulary list, similar to \cite{renPopMAGPopMusic2020}. Up to three tracks are used in this work: ``track\_0'' is the melody track, ``track\_1'' is the bass track, and ``track\_2'' is an accompaniment track. The track token is the first token of all the tokens in that track. More tracks can be added in the future if they are arranged in the same order, e.g. track\_3 for drum and track\_4 for a second melody.

\begin{figure}
 \centerline{\framebox{
 \includegraphics[width=0.7\columnwidth]{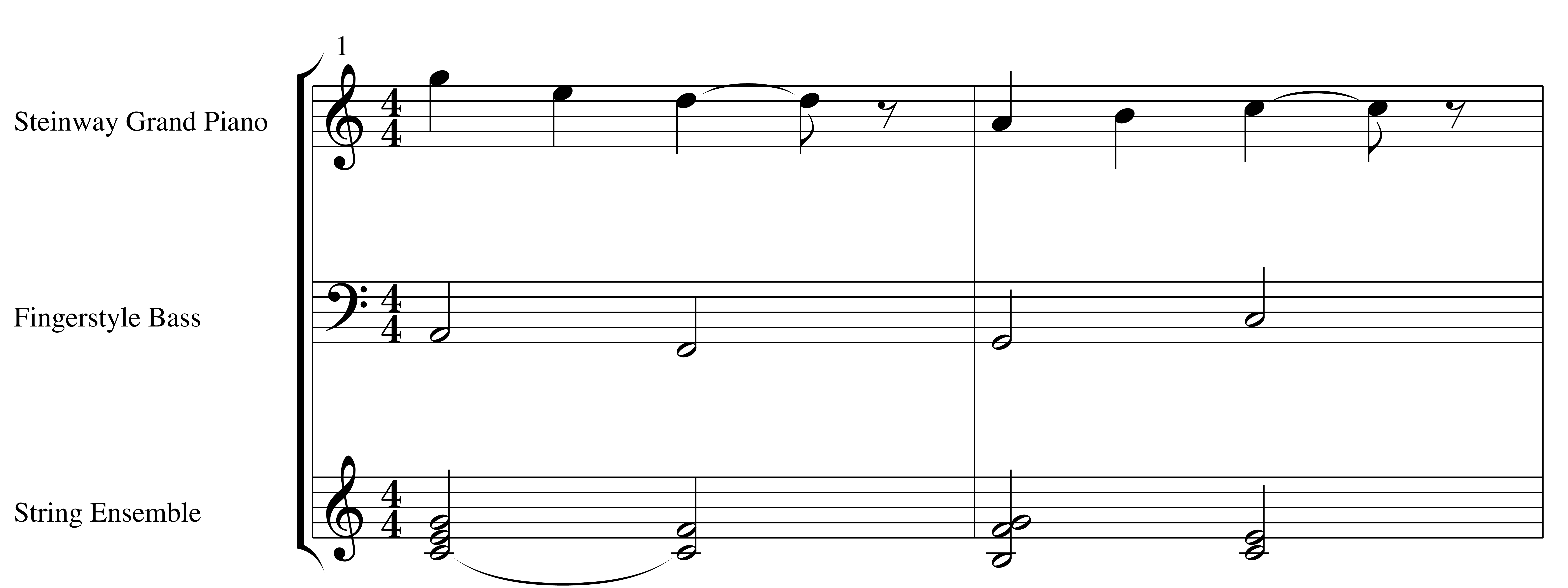}}}
 \caption{Example of a musical segment in our dataset.}
 \label{fig:example_sheet}
\end{figure}

\figurename ~\ref{fig:example_sheet} shows a piece with three tracks. Before the calculated control information is added, the event list is: \url{4/4, t_3, i_0, i_32, i_48, bar, track_0, e_0,p_79, n_4, e_4, p_76, n_4, e_8, p_74, n_6, track_1, e_0, p_45, n_8, e_8, p_41, n_8, track_2, e_0, p_64, p_67, n_8, e_0, p_60, n_16, e_8, p_65, n_8, bar, track_0, e_0, p_69, n_4, e_4, p_71, n_4, e_8, p_72, n_6, track_1, e_0, p_43, n_8, e_8, p_48, n_8, track_2, e_0, p_59, p_65, p_67, n_8, e_8, p_60, p_64, n_8}.

The control information that is included in our proposed framework is tensile strain (12 categories), cloud diameter (12 categories), track density/ polyphony/occupation rate (each for 10 categories) as per the previous subsection. Because the calculation of the bar tonal tension is based on a determined key, the key of the song is also determined and added to the music input. After those calculated control tokens are added, the data representation for \figurename ~\ref{fig:example_sheet} becomes: \url{4/4, t_3, k_0, d_0, d_0, d_0, o_8, o_9, o_9,  y_0, y_0, y_9, i_0, i_32, i_48, bar, s_2, a_1, track_0, e_0, p_79, n_4, e_4, p_76, n_4, e_8, p_74, n_6, track_1, e_0, p_45, n_8, e_8, p_41, n_8, track_2, e_0, p_64, p_67, n_8, e_0, p_60, n_16, e_8, p_65, n_8, bar, s_5, a_6, track_0, e_0, p_69, n_4, e_4, p_71, n_4, e_8, p_72, n_6, track_1, e_0, p_43, n_8, e_8, p_48, n_8, track_2, e_0, p_59, p_65, p_67, n_8, e_8, p_60, p_64, n_8}. The tokens at the start of the event list are time signature, tempo, and key tokens. The track control tokens appear after the key token, followed by the instrument tokens. A ``bar'' token follows the instrument token, immediately followed by tension control. The ``track'' token is followed by the ``position'', ``pitch'' and ``duration'' tokens inside each track. The final vocabulary list is represented in \tablename~\ref{tab:all_vocab}. 

\begin{table}[]
\centering
\caption{The event vocabulary, including all calculated control tokens.}
\label{tab:all_vocab}
\begin{tabular}{@{}lll@{}}
\toprule
token types      & tokens                                            & number \\ \midrule
position         & e\_0...e\_15                                      & 16     \\
pitch            & p\_21...p\_108                                    & 88     \\
duration         & n\_1...n\_32                                      & 32     \\
structure tokens & bar, track\_0, track\_1, track\_2 & 4      \\
time signature   & 4/4, 3/4, 2/4, 6/8                                & 4      \\
tempo            & t\_0...t\_6                                       & 7      \\
instrument       & i\_0...i\_127                                     & 128    \\
key              & k\_0...k\_23                                      & 24     \\
tensile strain   & s\_0...s\_11                                      & 12     \\
cloud diameter   & a\_0...a\_11                                      & 12     \\
density          & d\_0...d\_9                                       & 10     \\
polyphony        & y\_0...y\_9                                       & 10     \\
occupation       & y\_0...o\_9                                       & 10     \\
model            & mask, pad, eos                                    & 3      \\
total            &                                                   & 360    \\ \bottomrule
\end{tabular}
\end{table}

\subsection{Model architecture}

As the core task here is music infilling rather than forward generation, the model should ideally use bidirectional information. The transformer encoder-decoder model \cite{vaswani2017attention} which was originally developed for the seq-seq translation task, is adapted in this work. 
The infilling task in music can be likened to the corrupted token reconstruction task in natural language processing\cite{devlin2018bert}. In our proposed framework, a transformer encoder-decoder is used to reconstruct the masked input in the encoder\cite{song2019mass}. The bi-directional encoder makes each token in the encoder attend to other positions in the input, while the token in a one stack decoder language model can only attend to the tokens before the current token\cite{raffelExploringLimitsTransfer2020}. 
Our model has the same structure as the vanilla transformer \cite{vaswani2017attention} with two stages of training. Firstly, music grammar is learned in the pretraining stage and then specific tasks are learned in the finetuning stage. This process is similar to the work of \cite{devlin2018bert,zeng2021musicbert,chou2021MIDIbert}.

During pretraining, we accustom the model to small masked sections: one ``mask'' token can replace up to three tokens. If the input $x$ position from $u$ to $v$ is masked, and the $l=u-v$ is the masked token span length, the loss function is calculated as in Eq.~\eqref{loss}:

\begin{equation}\label{loss}
  L(\theta) = \log P(x^{u:v}|x^{\backslash u:v};\theta),  ~~ 0<u-v<=3.
\end{equation}

Up to 15\% of the tokens in the input are randomly masked with a ``mask'' in pretraining. We only use one ``mask'' token to replace each span, which differs from other work \cite{raffelExploringLimitsTransfer2020} which uses a different mask token for each span masked. The lengths of the spans of the masked token are 3, 1, 2 and the frequency of the masked tokens with those span lengths is in the ratio of 2:1:1 in the training respectively.

After pretraining, the finetuning stage is used to train the model for the real application task with larger masked areas). The finetuning task includes three masking types corresponding to the application. For each song: 
1. randomly select a bar, and mask all tracks in that bar. 
2. randomly select a track, and mask all the bars in selected tracks. 
3. randomly select bars, and randomly select tracks in that bar.

\begin{figure}
\includegraphics[width=\textwidth]{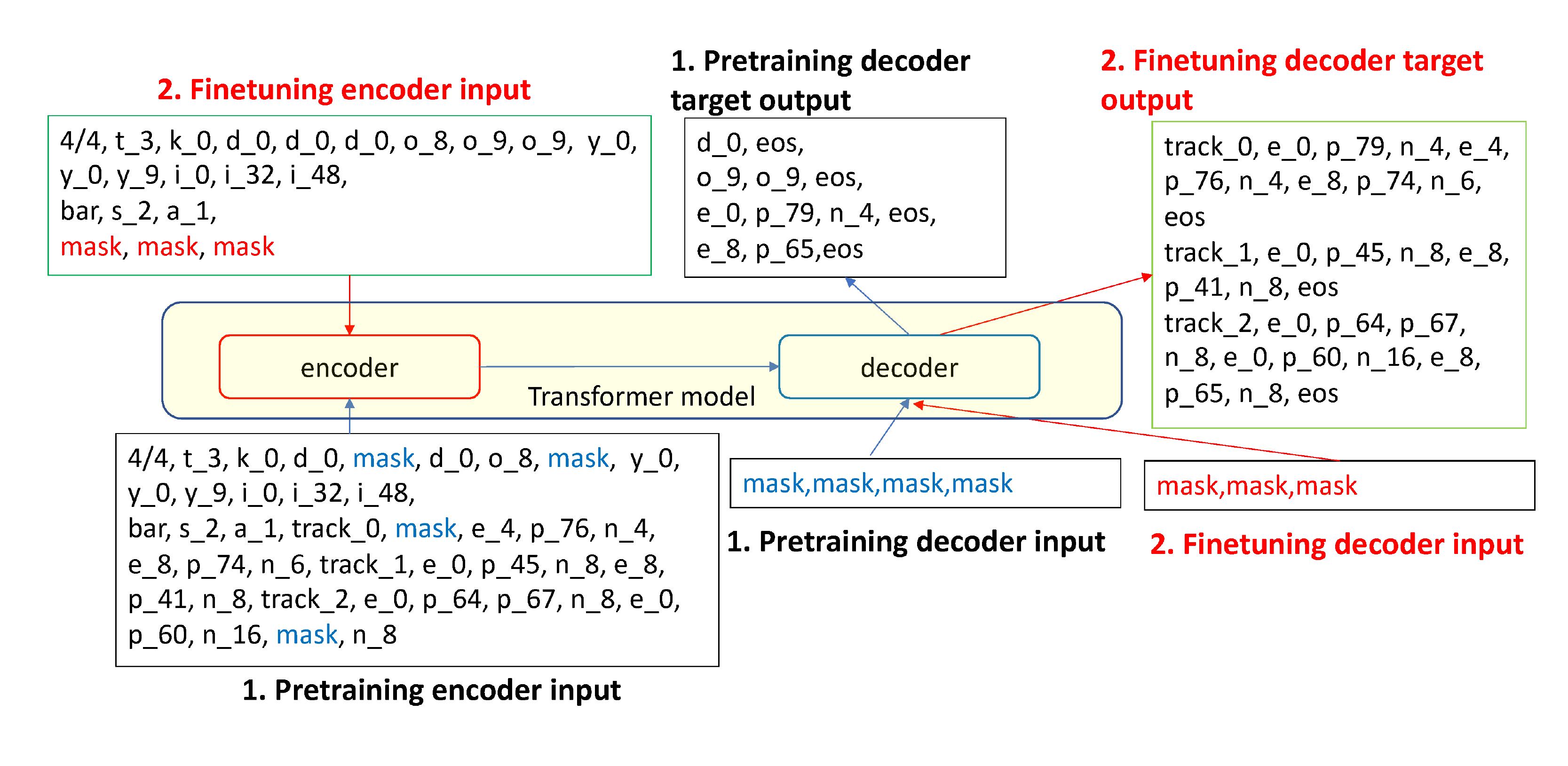}
\caption{The model encoder input, decoder input and decoder target output during pretraining and finetuning. The maximum masked span length is three for pretraining, and a ``mask'' token will replace a track in a bar during the finetuning stage.} 
\label{fig:modelinput}
\end{figure}

One ``mask'' token represents a track in a bar, and the decoder target is to reconstruct that masked bar track. Each ``mask'' in the encoder input is matched with a ``mask'' input in the decoder, and the decoder target output will end with an ``eos'' token. A ``pad'' token is also added to pad sequences of different lengths to match the batch size. \figurename ~\ref{fig:modelinput} shows masked encoder input and the decoder input and target output during pretraining/finetuning. During finetuning, if the first bar of \figurename ~\ref{fig:example_sheet} is infilled, the encoder input becomes: \url{4/4, t_3, k_0, d_0, d_0, d_0, o_8, o_9, o_9,  y_0, y_0, y_9, i_0, i_32, i_48, bar, s_2, a_1, mask, mask, mask, bar, s_5, a_6, track_0, e_0, p_69, n_4, e_4, p_71, n_4, e_8, p_72, n_6, track_1, e_0, p_43, n_8, e_8, p_48, n_8, track_2, e_0, p_59, p_65, p_67, n_8, e_8, p_60, p_64, n_8}. The decoder input is:\url{mask,mask,mask}, and the decoder target output is \url{track_0, e_0, p_79, n_4, e_4, p_76, n_4, e_8, p_74, n_6, eos, track_1, e_0, p_45, n_8, e_8, p_41, n_8, eos, track_2, e_0, p_64, p_67, n_8, e_0, p_60, n_16, e_8, p_65, n_8, eos}. We omitted the second bar's tokens to save page space.

\section{Experimental setup}

We conducted an experiment to validate the musical quality of the output as well as the influence of the control features. Two models with the same vocabulary size were trained in the experiment: one with controls and one without. The model without control will not add the six calculated controls to the input.

\subsection{Dataset}

Any dataset used by our proposed model should have proper track/bar numbers. The maximum number of tracks in a song is limited to three, which includes mandatory melody and bass tracks, and an optional accompaniment track. The maximum bar length in a song is set to 16, which is enough for the infilling reconstruction and not too long for the model to compute.

To build our symbolic dataset, we filter the Lakh LMD-matched dataset \cite{raffel2016learning} for songs that have both a melody and bass track, as well as an optional accompaniment track. After that, the key of the song and the tension metrics are calculated using \cite{cuthbert2010music21,guoMIDIMinerPython2019}. A total of 32,352 songs remain after this step. To tackle the imbalance of the time signature in the remaining files, a subset with metre 2/4, 3/4 and 6/8 is pitch-shifted to the other 11 keys in the same mode. The same strategy is applied to a subset of songs with minor keys. A moving window size of 8 bars is used to create multiple dataset entries from a single song. All the calculated control features are added in this step.

\subsection{Model configuration and training}\label{control_list}
One model is trained with with all input and control tokens, the second models is trained without adding the control tokens. Both the encoder and decoder of the transformer have 4 layers, 8 heads, and the model dimension is 512. Both the models are trained for 10 epochs with 2 epochs of pretraining and the remaining 8 epochs for fine-tuning. The learning rate is 0.0001, and the training takes around 2 days per model on a Tesla V100 GPU. The training/validation/test data split ratio is 8:1:1.

\subsection{Inference strategy}

Our token representation allows us to guide the model to only generate output that adheres to the representation's grammar. The grammar of the notes in a track in the regular expression format is (step pitch $+$ duration)$*$. In the inference stage, the tokens not in this grammar are removed by setting those notes' logit to -100, and then weighted sampling is applied to sample from the tokens. This makes sure that the output will not sample the incorrect tokens and the result always has a correct grammar.

\section{Evaluation}

To evaluate the generated music infillings generated by the model with and without controls, we select seven objective metrics based on pitch and rhythm similarity. We compare the difference of those features between the generated and the original music in the masked position. Then we check if our model can really control features of the generated music by changing track/bar controls through our developed Google Colab interface. Our experiment evaluates if the generated music follows the desired control features and is musically meaningful.

\subsection{Objective evaluation using selected metrics}

To compare the quality of the generated infilling by those two models, we selected five pitch-related metrics and two duration-related metrics inspired by \cite{yang2020evaluation}. The infilling generation task makes it meaningful to compare the metrics' difference between the generated infilling and the original music. A smaller difference means the generated infilling has more stylistic similarity to the original music. Note that there is not only one optimal way to infill music, and we assume the original one is the target here. In future work, this assumption may be tested by allowing for a human listening experiment to evaluate the generated infillings. Both the track and bar infilling are evaluated. 

We selected 1,000 songs randomly from the testset and masked a random track/bar, to test each of the two models. The models then generate the infilling for the masked track/bar. Seven objective metrics are selected inspired by \cite{yang2020evaluation} including five pitch related metrics: 1) pitch number: the number of used pitches. 2) note number: the number of used notes. 3) pitch range: $pitch_{max} - pitch_{min}$. 4) chromagram histogram: the histogram of 12 pitch groups after categorising all the pitches 5) pitch interval histogram:  the histogram of the pitch difference between two consecutive notes. Two duration features: 6) duration histogram. 7) onset interval histogram: the histogram of time between two consecutive notes. These seven features are calculated for the generated/original infilled track/bar. For the first three features we calculate the absolute difference between the feature for the generated and original music, normalised by the feature of the original music: $abs(feature_{gen} - feature_{ori}) / feature_{ori}$ For the last four histogram features we calculate the sum of the square difference between the features of the generated and the original music, normalised by the sum of the square of the feature of the original music: $sum(square(feature_{gen} - feature_{ori})) / sum(square(feature_{ori}))$. 

The mean and the standard deviation are calculated on those difference features and  reported in \tablename~\ref{tab:obj}. The left value in each cell is the result for the model without added control tokens, and the right value is the result for the model with added control tokens. All of the values, except the track pitch number standard deviation, show that the model with added control generates music more similar to the original music, especially in terms of melody, accompaniment track, and bar infilling. The added control work much like a template, and the generated music follows these conditions well.

\begin{table}[]
\centering
\caption{The mean and standard deviation of the difference for the seven objective metrics between the generated and original music. The left value in each cell is the result from the model without added control tokens, and the right value is the result from the model with added control. The column header shows was was infilled: melody track, bass track,  accompaniment track, or a random bar (all tracks in this bar). }
\label{tab:obj} \footnotesize
\begin{tabular}{|c|cc|cc|cc|ll|}
\hline
 &
  \multicolumn{2}{c|}{Melody} &
  \multicolumn{2}{c|}{Bass} &
  \multicolumn{2}{c|}{Accompaniment} &
  \multicolumn{2}{c|}{Bar} \\ \hline
  features &
  \multicolumn{1}{c|}{mean} &
  std &
  \multicolumn{1}{c|}{mean} &
  std &
  \multicolumn{1}{c|}{mean} &
  std &
  \multicolumn{1}{c|}{mean} &
  \multicolumn{1}{c|}{std} \\ \hline
pitch number &
  \multicolumn{1}{c|}{0.45/0.39} &
  0.57/0.46 &
  \multicolumn{1}{c|}{0.55/0.52} &
  0.76/0.78 &
  \multicolumn{1}{c|}{0.57/0.41} &
  0.69/0.52 &
  \multicolumn{1}{l|}{0.41/0.33} &
  0.79/0.58 \\ \hline
note number &
  \multicolumn{1}{c|}{0.82/0.29} &
  2.71/0.85 &
  \multicolumn{1}{c|}{0.63/0.29} &
  1.25/0.71 &
  \multicolumn{1}{c|}{0.97/0.41} &
  2.41/0.73 &
  \multicolumn{1}{l|}{0.42/0.32} &
  0.90/0.71 \\ \hline
pitch range &
  \multicolumn{1}{c|}{0.59/0.49} &
  0.93/0.74 &
  \multicolumn{1}{c|}{0.62/0.58} &
  0.92/0.87 &
  \multicolumn{1}{c|}{0.80/0.55} &
  1.05/0.76 &
  \multicolumn{1}{l|}{0.55/0.38} &
  2.55/1.54 \\ \hline
chroma hist &
  \multicolumn{1}{c|}{0.59/0.53} &
  0.49/0.42 &
  \multicolumn{1}{c|}{0.45/0.38} &
  0.50/0.44 &
  \multicolumn{1}{c|}{0.34/0.27} &
  0.42/0.31 &
  \multicolumn{1}{l|}{0.73/0.58} &
  0.79/0.70 \\ \hline
pitch itv hist &
  \multicolumn{1}{c|}{0.44/0.39} &
  0.46/0.40 &
  \multicolumn{1}{c|}{0.50/0.45} &
  0.55/0.53 &
  \multicolumn{1}{c|}{1/0.84} &
  0.75/0.63 &
  \multicolumn{1}{l|}{0.66/0.60} &
  0.77/0.75 \\ \hline
duration hist &
  \multicolumn{1}{c|}{0.55/0.42} &
  0.61/0.46 &
  \multicolumn{1}{c|}{0.77/0.61} &
  0.77/0.68 &
  \multicolumn{1}{c|}{1.03/0.74} &
  0.93/0.78 &
  \multicolumn{1}{l|}{0.52/0.45} &
  0.64/0.61 \\ \hline
onset itv hist &
  \multicolumn{1}{c|}{0.45/0.35} &
  0.55/0.41 &
  \multicolumn{1}{c|}{0.74/0.61} &
  0.71/0.70 &
  \multicolumn{1}{c|}{0.86/0.69} &
  0.85/0.77 &
  \multicolumn{1}{l|}{0.44/0.38} &
  0.63/0.58 \\ \hline
\end{tabular}
\end{table}

\subsection{The interactive interface and controllability}

A Google Colab notebook has been prepared for the exploration of this application \footnote{\url{https://github.com/ruiguo-bio/MusIAC}}. The user can upload MIDI files or select MIDI files from the test dataset. Here  ``Imagine'' from John Lennon is selected from the test dataset as an example.

\begin{figure}[!ht]
\includegraphics[width=.7\textwidth]{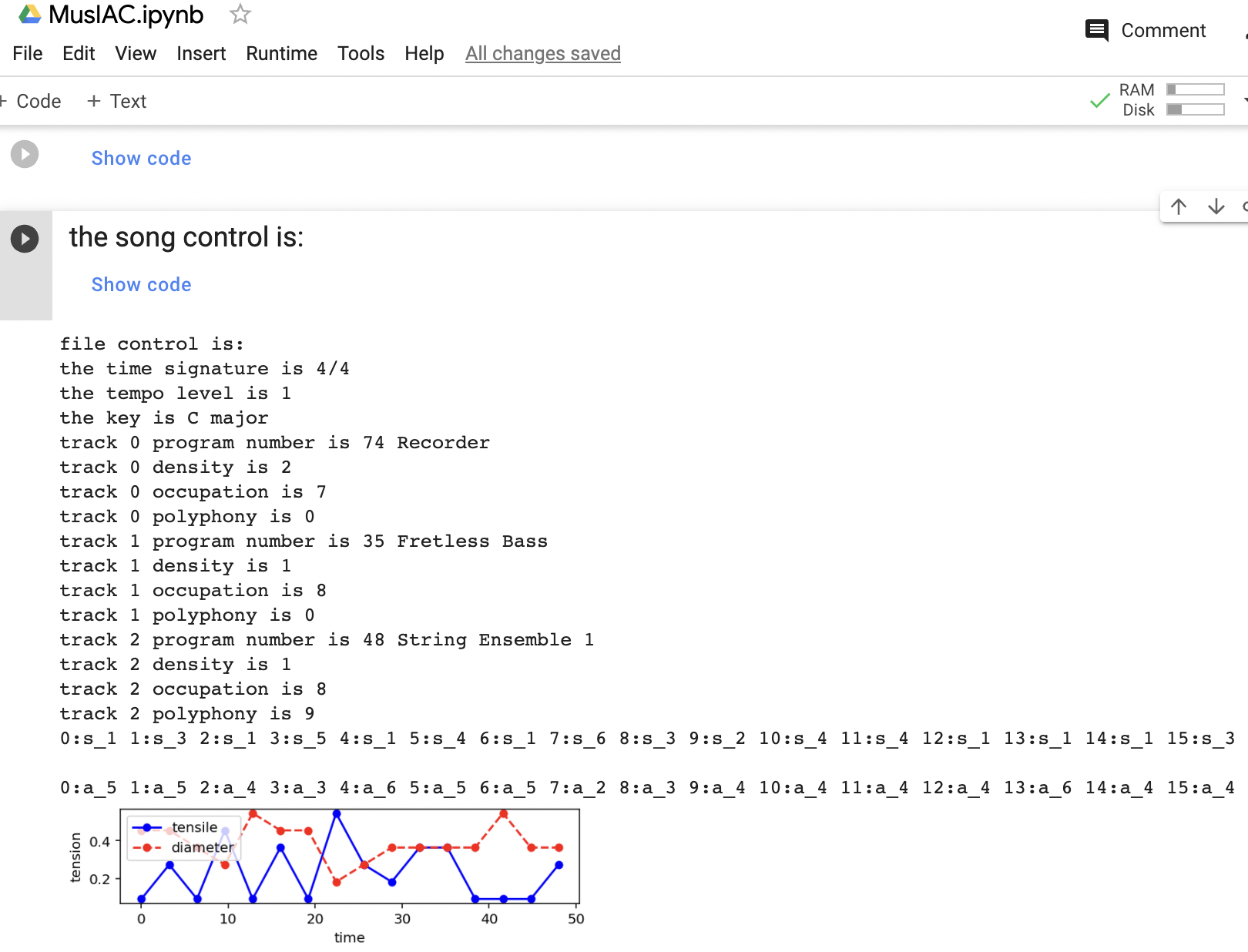}
\caption{The original music's control information including track/bar control and song key, tempo and time signature.} 
\label{fig:original}
\end{figure}

\begin{figure}[!ht]
\includegraphics[width=\textwidth]{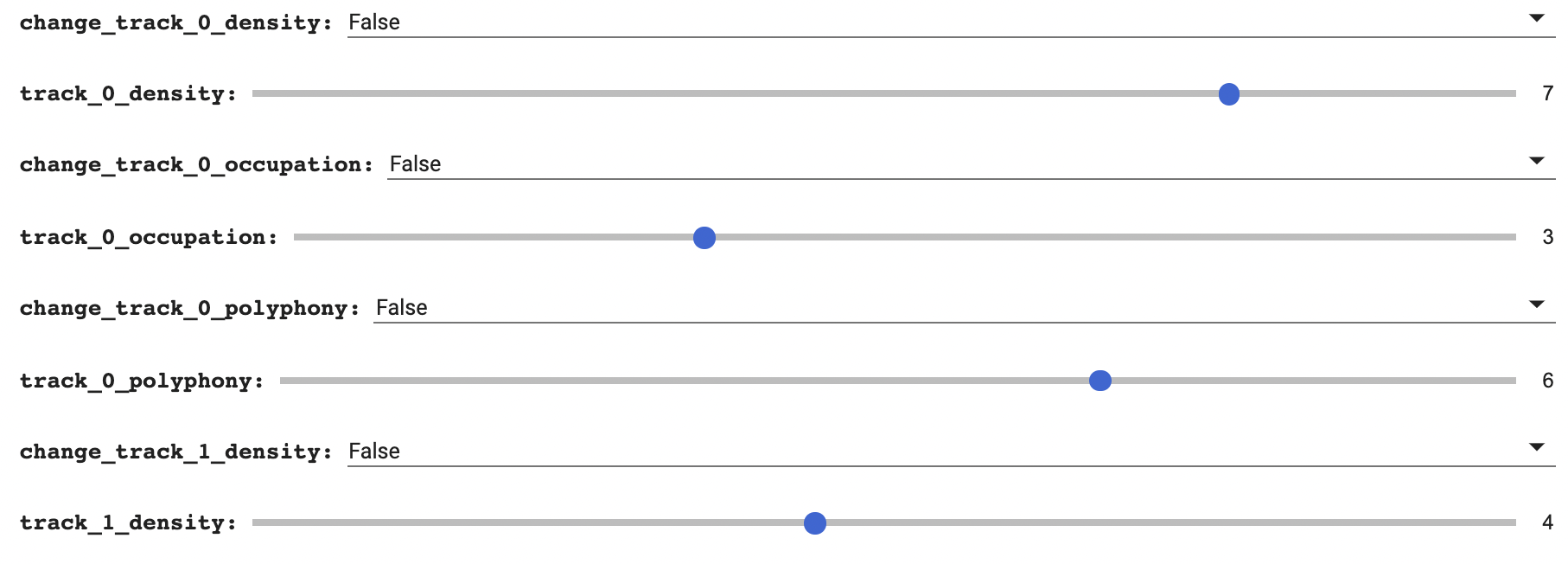}
\caption{The track/bar controls can be changed separately} 
\label{fig:control}
\end{figure}

\begin{figure}[!ht]
\centering
\begin{subfigure}{.5\textwidth}
  \centering
  \includegraphics[width=\linewidth]{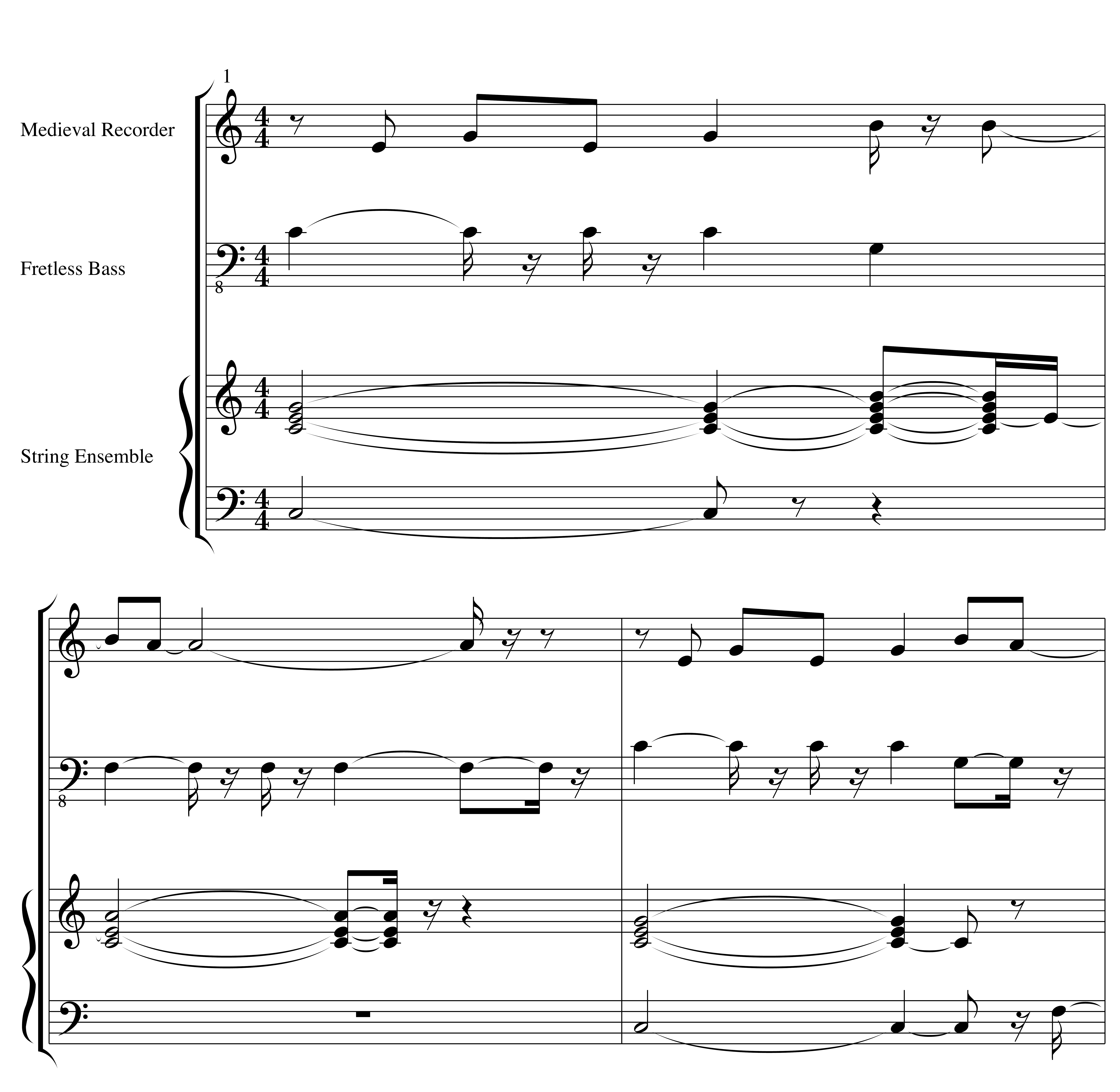}
  \label{fig:original_sheet}
\end{subfigure}%
\begin{subfigure}{.5\textwidth}
  \centering
  \includegraphics[width=\linewidth]{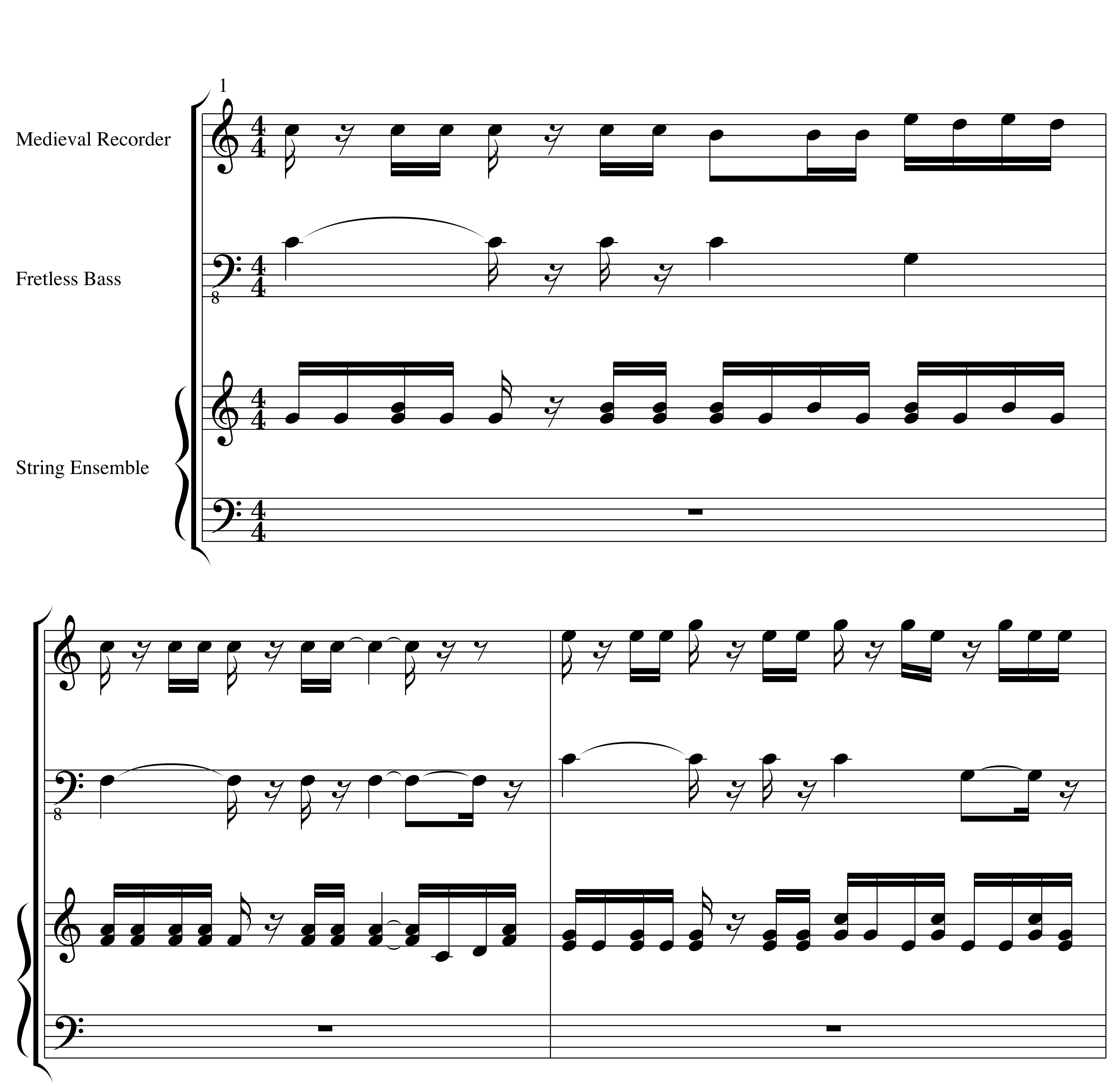}
  \label{fig:changed_sheet}
\end{subfigure}
\caption{The first three bars of a section of ``Imagine''. The left figure is the original, and the right figure shows the infilled melody and accompaniment track with changed track density level from 1 to 5 and polyphony rate from 9 to 2. The original melody track has a low note density level of 1. The accompaniment track has low note density level 1 and high polyphony level 9. The infilled melody/accompaniment track match the selected controls, and the accompaniment is similar to Alberti bass, with more notes and less polyphony. }
\label{fig:track_infilling}
\end{figure}

After selecting/uploading the song, the user can choose to infill a track/bar without changing the control tokens, or change the track/bar controls first, and then generate the corresponding track/bar. The original control tokens of one section of ``Imagine'' are calculated and shown as in the left figure of \figurename~\ref{fig:track_infilling}. The melody and accompaniment tracks have low note density, which means there are not many note onsets in that track. The accompaniment track is composed of mainly chord notes. The track/bar control can be changed by selecting the specific control type and value as shown in \figurename~\ref{fig:control}(only a section is shown due to page limitations). To add more notes to those tracks, and make the accompaniment track less polyphonic, we first change the melody track density to level 8 from level 1. After the new melody track is generated, the accompaniment track is generated with density set to level 5 from level 1 and the polyphony level set to level 2 from level 9. The generated result is shown in the right figure in \figurename~\ref{fig:track_infilling}. The resulting music matches the desired control with a tolerance threshold of 1(which means level 3 is accepted if the target level is 4 or 2). The resulting accompaniment track's texture is similar to Alberti bass, and both of the tracks have more notes added following the increase of the track density level control.

\begin{figure}[!ht]
\includegraphics[width=\textwidth]{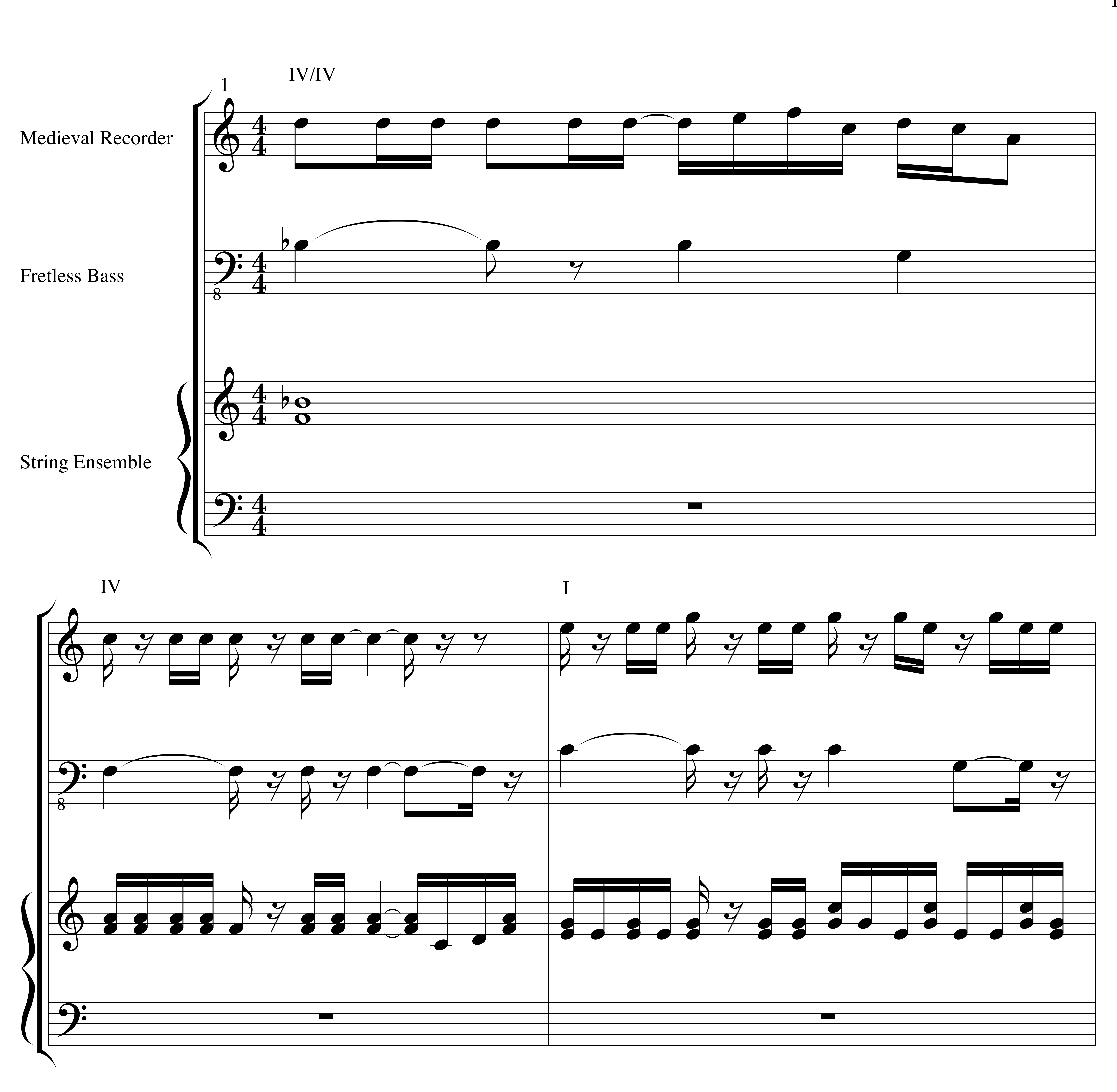}
\caption{The first bar tonal tension is changed from 1 to 6. Here the ``tensile strain'' is changed, and the result shows that the first bar is the subdominant of the IV chord of C major. The second bar is subdominant and goes to I in the third bar. This result increases the tension but also progresses smoothly with the surrounding of the music.} 
\label{fig:generated_bar}
\end{figure}

Based on the previous track's infilling result, the first bar's tensile strain is changed from level 1 to level 6 to increase the tension of the beginning. The infilled result is shown in \figurename~\ref{fig:generated_bar}. The first bar contains the subdominant of the F major chord, which is the second bar's chord. This new first bar, together with the following two bars gives the progression of \url{IV/IV -> IV -> I}, which is musically meaningful (from subdominant to tonic chord), and it also increases the tension of the first bar. The full 16 bars music/sheet of the original/generated music are in the supplement material.

The track/bar infilling operation can be repeated several times until the result is satisfactory. The generated MIDI/rendered wav file can be downloaded for further editing or imported in a digital audio workstation (DAW).

\section{Conclusion}

In this work, we propose an pretraining-finetuning transformer framework for the music infilling task with multiple levels of control, together with an intuitive interface. We selected track density, polyphony, and occupation level as track level controls to increase the user's controllability. This offers a greater steerability compared to existing systems with just density control\cite{ensMMMExploringConditional2020,hadjeres2021piano}. We also added tensile strain and cloud diameter features per bar as controls for the tonality (tonal tension) of each bar. Control tokens work as a template on which the generated music can be conditioned. The generated result from the input with those added controls as conditions has a higher stylistic similarity to the original music, versus a model without controls.

To optimally demonstrate our proposed framework with a user-friendly interactive interface, we have made it available through Google Colab. In this interface, the user can modify the music while it is being generated. 

In the future work, we will systematically check the controllability of each of the six control tokens and further evaluate the musicality with quantitative metrics. A listening test would also be useful to evaluate the musical quality, as there may be more good sounding possibilities than just the original music. We would also like to explore how to further increase the controllability of this model. Currently, our model learns to follow controls (i.e., features) that are already present or easy to calculate from our dataset. It is hard for the model to generate music with ``unseen'' musical features, i.e. hard to capture, implicit characteristics. In recent research, a transformer model was combined with a prior model to model the latent space \cite{taketo_akama2021}. If different music properties can be disentangled in the latent space\cite{ashis_pati2021}, this will allow for direct manipulation of the generated music's properties even though these features were not explicit in the dataset. 

\section{Acknowledgement}
This work is funded by Chinese scholarship Council and Singapore Ministry of Education Grant no. MOE2018-T2-2-161.



%
%
%

\bibliographystyle{splncs04}
\bibliography{guo}

\end{document}